%% file: main.tex
\documentclass[runningheads]{llncs}

\usepackage{amsfonts} 
\usepackage{amsmath}
\usepackage{amssymb}

\usepackage{amsthm}
\usepackage{bm}
\usepackage{booktabs}
\usepackage{cite}
\usepackage{color, colortbl}
\usepackage[inline]{enumitem}
\usepackage[T1]{fontenc}
\usepackage{graphicx}
\usepackage[utf8]{inputenc}
\usepackage{float}
\usepackage[symbol]{footmisc}
\usepackage{latexsym}
\usepackage{mathtools}
\usepackage{microtype}
\usepackage{multirow}
\usepackage{rotating}
\usepackage{tabularx}
\usepackage[hyphens]{url}
\usepackage{xspace}
\usepackage[colorlinks,breaklinks=true,bookmarks=false]{hyperref}
\hypersetup{linkcolor=black}
\input{notation}
\usepackage{authblk}

\definecolor{Gray}{gray}{0.97}

\makeatletter 
\newcommand{\printfnsymbol}[1]{%
  \textsuperscript{\@fnsymbol{#1}}%
}
\makeatother

\begin{document}
\pagestyle{headings}
\mainmatter
\def\ECCVSubNumber{6496}  

\title{A unifying mutual information view of \\ metric learning: cross-entropy vs. pairwise losses}
\titlerunning{Metric learning: cross-entropy vs. pairwise losses}

 \author{Malik~Boudiaf\thanks{Equal contributions}\inst{1}\orcidID{0000-0003-2047-6447} \and
 Jérôme~Rony\printfnsymbol{1}\inst{1}\orcidID{0000-0002-6359-6142} \and
 Imtiaz~Masud~Ziko\printfnsymbol{1}\inst{1}\orcidID{0000-0002-9254-2426} \and
 Eric~Granger\inst{1}\orcidID{0000-0001-6116-7945}\and
 Marco~Pedersoli\inst{1}\orcidID{0000-0002-7601-8640}\and
 Pablo~Piantanida\inst{2}\orcidID{0000-0002-8717-2117}\and
 Ismail~Ben~Ayed\inst{1}\orcidID{0000-0002-9668-8027}
 }
\author{Malik~Boudiaf\thanks{Equal contributions}\inst{1}\and
Jérôme~Rony\printfnsymbol{1}\inst{1}\and
Imtiaz~Masud~Ziko\printfnsymbol{1}\inst{1}\and
Eric~Granger\inst{1}\and
Marco~Pedersoli\inst{1}\and
Pablo~Piantanida\inst{2}\and
Ismail~Ben~Ayed\inst{1}
}
\authorrunning{M. Boudiaf et al.}
%
\institute{Laboratoire d'Imagerie, de Vision et d'Intelligence Artificielle (LIVIA),\\ ÉTS Montreal, Canada\\ \email{\{malik.boudiaf.1, jerome.rony.1, imtiaz-masud.ziko.1\}@etsmtl.net}\and
Laboratoire des Signaux et Systèmes (L2S),\\ CentraleSupelec-CNRS-Université Paris-Saclay, France}
\maketitle

\input{0_abstract}
\input{1_introduction}
\input{2_view_of_mi}
\input{3_pairwise}
\input{4_crossentropy}
\input{5_0_experiments}
\input{6_conclusion}

\section{Acknowledgments}

    This project was supported by the NSERC (Discovery Grant RGPIN 2019-05954). This project has received funding from the European Union’s Horizon 2020 research and innovation program under the Marie Skłodowska-Curie grant agreement 792464.

\clearpage
\bibliographystyle{splncs04}
\bibliography{biblio}

\clearpage
\input{7_appendix}

\end{document}

%% file: notation.tex
\DeclareMathAlphabet{\mathalphm}{OMS}{cmsy}{m}{n}
\DeclareMathAlphabet{\mathalphb}{OMS}{cmsy}{b}{n}

\newcommand{\w}{\bm{\theta}}
\newcommand{\x}{\bm{x}} 
\newcommand{\cc}{\bm{c}} 
\newcommand{\z}{\bm{z}} 

\newcommand{\Dcos}{D^{\text{cos}}}

\newcommand{\ceq}{\stackrel{\mathclap{\normalfont\mbox{c}}}{=}}
\newcommand{\capprox}{\stackrel{\mathclap{\normalfont\mbox{c}}}{\approx}}
\newcommand{\cleq}{\stackrel{\mathclap{\normalfont\mbox{c}}}{\leq}}
\newcommand{\cgeq}{\stackrel{\mathclap{\normalfont\mbox{c}}}{\geq}}

\newcommand{\ent}{\mathcal{H}}

\newcommand{\tran}{^\mathsf{T}}

\newcommand{\norm}[1]{\left\lVert#1\right\rVert}

\newcolumntype{L}[1]{>{\raggedright\let\newline\\\arraybackslash\hspace{0pt}}m{#1}}
\newcolumntype{C}[1]{>{\centering\let\newline\\\arraybackslash\hspace{0pt}}m{#1}}
\newcolumntype{R}[1]{>{\raggedleft\let\newline\\\arraybackslash\hspace{0pt}}m{#1}}

\makeatletter
\DeclareRobustCommand\onedot{\futurelet\@let@token\@onedot}
\def\@onedot{\ifx\@let@token.\else.\null\fi\xspace}

\def\eg{\emph{e.g}\onedot} 
\def\ie{\emph{i.e}\onedot}

\def\wrt{w.r.t\onedot} 

\makeatother

\DeclareMathOperator*{\argmin}{arg\,min}

\makeatletter

\newcommand\Autoref[1]{\@first@ref#1,@}
\def\@throw@dot#1.#2@{#1}
\def\@set@refname#1{
    \edef\@tmp{\getrefbykeydefault{#1}{anchor}{}}%
    \xdef\@tmp{\expandafter\@throw@dot\@tmp.@}%
    \ltx@IfUndefined{\@tmp autorefnameplural}%
         {\def\@refname{\@nameuse{\@tmp autorefname}s}}%
         {\def\@refname{\@nameuse{\@tmp autorefnameplural}}}%
}
\def\@first@ref#1,#2{%
  \ifx#2@\autoref{#1}\let\@nextref\@gobble
  \else%
    \@set@refname{#1}
    \@refname~\ref{#1}
    \let\@nextref\@next@ref
  \fi%
  \@nextref#2%
}
\def\@next@ref#1,#2{%
   \ifx#2@ and~\ref{#1}\let\@nextref\@gobble
   \else, \ref{#1}
   \fi%
   \@nextref#2%
}

\makeatother

%% file: 0_abstract.tex
\begin{abstract}

Recently, substantial research efforts in Deep Metric Learning (DML) focused on designing complex pairwise-distance losses, which require convoluted schemes to ease optimization, such as sample mining or pair weighting.
The standard cross-entropy loss for classification has been largely overlooked in DML.
On the surface, the cross-entropy may seem unrelated and irrelevant to metric learning as it does not explicitly involve pairwise distances.
However, we provide a theoretical analysis that links the cross-entropy to several well-known and recent pairwise losses.
Our connections are drawn from two different perspectives: one based on an explicit optimization insight; the other on discriminative and generative views of the mutual information between the labels and the learned features.
First, we explicitly demonstrate that the cross-entropy is an upper bound on a new pairwise loss, which has a structure similar to various pairwise losses: it minimizes intra-class distances while maximizing inter-class distances.
As a result, minimizing the cross-entropy can be seen as an approximate \emph{bound-optimization} (or \emph{Majorize-Minimize}) algorithm for minimizing this pairwise loss.
Second, we show that, more generally, minimizing the cross-entropy is actually equivalent to maximizing the mutual information, to which we connect several well-known pairwise losses. 
Furthermore, we show that various standard pairwise losses can be explicitly related to one another via bound relationships.
Our findings indicate that the cross-entropy represents a proxy for maximizing the mutual information -- as pairwise losses do -- without the need for convoluted sample-mining heuristics. Our experiments\footnote[2]{Code available at: \url{https://github.com/jeromerony/dml_cross_entropy}} over four standard DML benchmarks 
strongly support our findings. We obtain state-of-the-art results, outperforming recent and complex DML methods. \keywords{Metric Learning, Deep Learning, Information Theory}
\end{abstract}

%% file: 1_introduction.tex
\section{Introduction}
The core task of metric learning consists in learning a metric from high-dimensional data, such that the distance between two points, as measured by this metric, reflects their semantic similarity. Applications of metric learning include image retrieval, zero-shot learning or person re-identification, among others. Initial attempts to tackle this problem tried to learn metrics directly on the input space \cite{lowe1995similarity}. Later, the idea of learning suitable embedding was introduced, with the goal of learning Mahalanobis distances \cite{xing2003distance, schultz2004learning, goldberger2005neighbourhood, weinberger2009distance, davis2007information}, which corresponds to learning the best linear projection of the input space onto a lower-dimensional manifold, and using the Euclidean distance as a metric. Building on the embedding-learning ideas, several papers proposed to learn more complex mappings, either by kernelization of already existing linear algorithms \cite{davis2007information}, or by using a more complex hypothesis such as linear combinations of gradient boosted regressions trees \cite{kedem2012non}.

The recent success of deep neural networks at learning complex,  nonlinear mappings of high-dimensional data aligns with the problem of learning a suitable embedding. Following works on Mahalanobis distance learning, most Deep Metric Learning (DML) approaches are based on pairwise distances. Specifically, the current paradigm is to learn a deep encoder 
that maps points with high semantic similarity
close to each other in the embedded space (\wrt pairwise Euclidean or cosine distances). 
This paradigm concretely translates into {\em pairwise losses} that encourage 
small distances for pairs of samples from the same class
and large distances for pairs of samples from different classes.
While such formulations seem intuitive, the practical implementations and optimization schemes for pairwise losses may become cumbersome, and randomly assembling pairs of samples typically results in slow convergence or degenerate solutions \cite{hermans2017defense}. Hence, research in DML focused on finding efficient ways to reformulate, generalize and/or improve sample mining and/or sample weighting strategies over the existing pairwise losses. Popular pairwise losses include triplet loss and its derivatives \cite{hermans2017defense, sohn2016improved, song2016deep, zheng2019hardness, ge2018deep}, contrastive loss and its derivatives \cite{hadsell2006dimensionality, wang2019multi}, Neighborhood Component Analysis and its derivatives \cite{goldberger2005neighbourhood, movshovitz2017no, wu2018improving}, among others. However, such modifications are often heuristic-based, and come at the price of increased complexity and additional hyper-parameters, reducing the potential of these methods in real-world applications. Furthermore, the recent experimental study in \cite{reality_check} showed that the improvement brought by an abundant metric learning literature in the last 15 years is at best marginal when the methods are compared fairly.

Admittedly, the objective of learning a useful embedding of data points intuitively aligns with the idea of directly acting on the distances between pairs of points in the embedded space. Therefore, the standard cross-entropy loss, widely used in classification tasks, has been largely overlooked by the DML community, most likely due to its apparent irrelevance for Metric Learning \cite{center_loss}. As a matter of fact, why would anyone use a point-wise prediction loss to enforce pairwise-distance properties on the embedding space? Even though the cross-entropy was shown to be competitive for face recognition applications \cite{deephypersphere, wang2018cosface, wang2018additivesoftmax}, to the best of our knowledge, only one paper empirically observed competitive results of a normalized, temperature-weighted version of the cross-entropy in the context of deep metric learning \cite{zhai2018classification}. However, the authors did not provide any theoretical insights for these results. 

On the surface, the standard cross-entropy loss may seem unrelated to the pairwise losses used in DML. Here, we provide theoretical justifications that connect directly the cross-entropy to several well-known and recent pairwise losses. Our connections are drawn from two different perspectives; one based on an explicit optimization insight and the other on mutual-information arguments. We show that four of the most prominent pairwise metric-learning losses, as well as the standard cross-entropy, are maximizing a common underlying objective: the Mutual Information (MI) between the learned embeddings and the corresponding samples' labels. As sketched in \autoref{sec:two_views_on_mi}, this connection can be intuitively understood by writing this MI in two different, but equivalent ways. Specifically, we establish tight links between pairwise losses and the \textit{generative} view of this MI. We study the particular case of contrastive loss \cite{hadsell2006dimensionality}, explicitly showing its relation to this MI. We further generalize this reasoning to other DML losses by uncovering tight relations with contrastive loss.
As for the cross-entropy, we demonstrate that the cross-entropy is an upper bound on an underlying pairwise loss -- on which the previous reasoning can be applied -- which has a structure similar to various existing pairwise losses. As a result, minimizing the cross-entropy can be seen as an approximate \emph{bound-optimization} (or \emph{Majorize-Minimize}) algorithm for minimizing this pairwise loss, implicitly minimizing intra-class distances and maximizing inter-class distances.  We also show that, more generally, minimizing the cross-entropy is equivalent to maximizing the \textit{discriminative} view of the mutual information. Our findings indicate that the cross-entropy represents a proxy for maximizing the mutual information, as pairwise losses do, without the need for complex sample-mining and optimization schemes. Our comprehensive experiments over four standard DML benchmarks (CUB200, Cars-196, Stanford Online Product and In-Shop) strongly support our findings. We consistently obtained state-of-the-art results, outperforming many recent and complex DML methods.

\subsubsection{Summary of contributions}
\begin{enumerate}
    \item Establishing relations between several pairwise DML losses and a generative view of the mutual information between the learned features and labels;
    \item Proving explicitly that optimizing the standard cross-entropy corresponds to an approximate bound-optimizer of an underlying pairwise loss;
    \item More generally, showing that minimizing the standard cross-entropy loss is equivalent to maximizing a discriminative view of the mutual information between the features and labels. 
    \item Demonstrating state-of-the-art results with cross-entropy on several DML benchmark datasets.
\end{enumerate}

%% file: 2_view_of_mi.tex
\section{On the two views of the mutual information}\label{sec:two_views_on_mi}

\begin{table}
    \caption{Definition of the random variables and information measures used in this paper.}
    \label{table:notations}
    \centering
    \begin{tabular}{lc}
    \multicolumn{2}{c}{General} \\
    \toprule
    Labeled dataset & $\mathcal{D}=\{(\x_i, y_i)\}_{i=1}^n$ \\
    \midrule
    Input feature space & $\mathcal{X}$ \\
    \midrule
    Embedded feature space & $\mathcal{Z} \subset \mathbb{R}^d$ \\
    \midrule
    Label/Prediction space & $\mathcal{Y} \subset \mathbb{R}^K$ \\
    \midrule
    Euclidean distance & $D_{ij} = \left\| \z_i - \z_j \right\|_2$ \\
    \midrule
    Cosine distance & $\Dcos_{ij} = \frac{\z_i\tran  \z_j}{\left\| \z_i \right\| \left\| \z_j \right\|}$ \\
    \bottomrule
    \end{tabular}
    \qquad
    \begin{tabular}{lc}
    \multicolumn{2}{c}{Model} \\
    \toprule
    Encoder & $\phi_\mathcal{W}: \mathcal{X} \rightarrow{\mathcal{Z}}$ \\
    \midrule
    Soft-classifier & $f_{\w} : \mathcal{Z} \rightarrow{[0, 1]^K}$ \\
    \bottomrule \\
    \multicolumn{2}{c}{Random variables (RVs)} \\
    \toprule
    Data &  $X$, $Y$ \\
    \midrule
    Embedding &  $\widehat{Z}|X \sim \phi_\mathcal{W}(X)$ \\
    \midrule 
    Prediction &  $\widehat{Y}|\widehat{Z} \sim f_{\w}(\widehat{Z})$ \\
    \bottomrule
    \end{tabular}
    \begin{tabular}{lc}
    \multicolumn{2}{c}{\rule{0pt}{4ex}Information measures} \\
    \toprule
    Entropy of $Y$ & $ \mathcal{H}({Y})\coloneqq \mathbb{E}_{p_{Y}}\left[-\log p_Y (Y)\right]$ \\
    \midrule
    Conditional entropy of $Y$ given $Z$  & $ \mathcal{H}({Y|\widehat{Z}})\coloneqq \mathbb{E}_{p_{Y\widehat{Z}}} \left[-\log {p}_{Y|\widehat{Z}} (Y|\widehat{Z})\right]$ \\
    \midrule
    Cross entropy (CE) between $Y$ and $\widehat{Y}$  & $\mathcal{H}(Y; \widehat{Y}) \coloneqq \mathbb{E}_{p_Y} \left[-\log p_{\widehat{Y}} (Y)\right]$ \\
    \midrule
    Conditional CE given $\widehat{Z}$  & $\mathcal{H}(Y; \widehat{Y}|\widehat{Z}) \coloneqq \mathbb{E}_{p_{\widehat{Z}Y}} \left[-\log p_{\widehat{Y}|\widehat{Z}} (Y|\widehat{Z})\right]$ \\
    \midrule
    Mutual information between $\widehat{Z}$ and $Y$ & $\mathcal{I}(\widehat{Z};Y) \coloneqq \mathcal{H}({Y}) - \mathcal{H}({Y}|\widehat{Z})$  \\
    \bottomrule
    \end{tabular}
\end{table}

The Mutual Information (MI) is a well known-measure designed to quantify the amount of information shared by two random variables. Its formal definition is presented in \autoref{table:notations}. Throughout this work, we will be particularly interested in $\mathcal{I}(\widehat{Z}; Y)$ which represents the MI between learned 
features $\widehat{Z}$ and labels $Y$. 
Due to its symmetry property, the MI can be written in two ways, which we will refer to as the \textit{discriminative view} and \textit{generative view} of MI:
\begin{equation}
    \mathcal{I}(\widehat{Z}; Y) = \underbrace{\ent(Y) - \ent(Y|\widehat{Z})}_{\text{discriminative view}} = \underbrace{\ent(\widehat{Z}) - \ent(\widehat{Z}|Y)}_{\text{generative view}}
\end{equation}
While being analytically equivalent, these two views present two different, complementary interpretations. In order to maximize $\mathcal{I}(\widehat{Z}; Y)$, the discriminative view conveys that the labels should be balanced (out of our control) and 
easily identified from the features. On the other hand, the generative view conveys that the features learned should spread as much as possible in the feature space, while keeping samples sharing the same class close to each other. Hence, the discriminative view is more focused on label identification, while the generative view focuses on more explicitly shaping the distribution of the features learned by the model. Therefore, the MI enables us to draw links between classification losses (\eg cross-entropy) and feature-shaping losses (including all the well-known pairwise metric learning losses).

%% file: 3_pairwise.tex
\section{Pairwise losses and the generative view of the MI}\label{sec:pairwise_losses}

In this section, we study four pairwise losses used in the DML community: center loss \cite{center_loss}, contrastive loss \cite{hadsell2006dimensionality}, Scalable Neighbor Component Analysis (SNCA) loss \cite{wu2018improving} and Multi-Similarity (MS) loss \cite{wang2019multi}. We show that these losses can be interpreted as proxies for maximizing the generative view of mutual information $\mathcal{I}(\widehat{Z};Y)$. We begin by analyzing the specific example of contrastive loss, establishing its tight link to the MI, and further generalize our analysis to the other pairwise losses (see \autoref{table:losses}). Furthermore, we show that these pairwise metric-learning losses can be explicitly linked to one another via bound relationships. 

\subsection{The example of contrastive loss}\label{subsec:contrastive_example}

We start by analyzing the representative example of contrastive loss \cite{hadsell2006dimensionality}. For a given margin $m \in \mathbb{R}^+$, this loss is formulated as:
\begin{equation}
    \mathcal{L}_\text{contrast} = \underbrace{\frac{1}{n} \sum_{i=1}^n \smashoperator[r]{\sum_{j:y_j=y_i}} D_{ij}^2}_{T_\text{contrast}} + \underbrace{\frac{1}{n} \sum_{i=1}^n \smashoperator[r]{\sum_{j:y_j\neq y_i}} [m - D_{ij}]_{+}^2}_{C_\text{contrast}}
\end{equation}
where $[x]_+=\max(0,x)$. This loss naturally breaks down into two terms: a \textit{tightness} part $T_\text{contrast}$ and a \textit{contrastive} part $C_\text{contrast}$. The tightness part encourages samples from the same class to be close to each other and form \textit{tight} clusters. As for the \textit{contrastive} part, it forces samples from different classes to stand far apart from one another in the embedded feature space. Let us analyze these two terms from a mutual-information perspective.

As shown in the next subsection, the tightness part of contrastive loss is equivalent to the tightness part of the center loss \cite{center_loss}: $T_\text{contrast} \ceq T_\text{center} = \frac{1}{2} \sum_{i=1}^n \left\| \z_i - \cc_{y_i} \right\|^2$, where $\cc_k = \frac{1}{|\mathcal{Z}_k|}\sum_{\z \in \mathcal{Z}_k} \z$ denotes the mean of feature points from class $k$ in embedding space $\mathcal{Z}$ and symbol $\ceq$ denotes equality up to a multiplicative and/or additive constant. Written in this way, we can interpret $T_\text{contrast}$ as a conditional cross entropy between $\widehat{Z}$ and another random variable $\Bar{Z}$, whose conditional distribution given $Y$ is a standard Gaussian centered around $\cc_Y$: $\Bar{Z}|Y \sim \mathcal{N}(c_Y, I)$:
\begin{equation}
    T_\text{contrast} \ceq \ent(\widehat{Z}; \Bar{Z}|Y) = \mathcal{H}(\widehat{Z}|Y) + \mathcal{D}_{KL}(\widehat{Z}||\Bar{Z}|Y)
\end{equation}
As such, $T_\text{contrast}$ is an upper bound on the conditional entropy that appears in the mutual information: 
\begin{equation}
    T_\text{contrast} \geq \mathcal{H}(\widehat{Z}|Y)
\end{equation}
This bound is tight when $\widehat{Z}|Y \sim \mathcal{N}(\cc_Y, I)$. Hence, minimizing $T_\text{contrast}$ can be seen as minimizing $\mathcal{H}(\widehat{Z}|Y)$, which exactly encourages the encoder $\phi_\mathcal{W}$ to produce low-entropy (=compact) clusters in the feature space for each given class. Notice that using this term only will inevitably lead to a trivial encoder that maps all data points in $\mathcal{X}$ to a single point in the embedded space $\mathcal{Z}$, hence achieving a global optimum.

To prevent such a trivial solution, a second term needs to be added. This second term -- that we refer to as the \textit{contrastive} term -- is designed to push each point away from points that have a different label. In this term, only pairs such that $D_{ij} \leq m$ produce a cost. Given a pair $(i,j)$, let us define $x=D_{ij}/m$. Given that $x \in [0,1]$, one can show the following: $1-2x \leq (1-x)^2 \leq 1-x$. Using linear approximation $(1-x)^2 \approx 1-2x$ (with error at most $x$), we obtain:
\begin{equation}\label{eq:contrastive_contrast}
    C_\text{contrast} \capprox -\frac{2m}{n} \sum_{i=1}^n \smashoperator[r]{\sum_{j:y_j\neq y_i}} D_{ij} = -\frac{2m}{n} \sum_{i=1}^n\sum_{j=1}^n D_{ij} + \frac{2m}{n} \sum_{i=1}^n \smashoperator[r]{\sum_{j:y_j=y_i}} D_{ij}
\end{equation}
While the second term in \autoref{eq:contrastive_contrast} is redundant with the tightness objective, the first term is close to the differential entropy estimator proposed in \cite{wang2011information}:
\begin{equation}\label{eq:entropy_estimator}
    \widehat{\ent}(\widehat{Z}) = \frac{d}{n(n-1)} \sum_{i=1}^n\sum_{j=1}^n \log D_{ij}^2 \ \ceq \ \sum_{i=1}^n\sum_{j=1}^n \log D_{ij}
\end{equation}
Both terms measure the spread of $\widehat{Z}$, even though they present different gradient dynamics.
All in all, minimizing the whole contrastive loss can be seen as a proxy for maximizing the MI between the labels $Y$ and the embedded features $\widehat{Z}$:
\begin{equation}
    \mathcal{L}_\text{contrast} = 
    \underbrace{\frac{1}{n}\sum_{i=1}^n\smashoperator[r]{\sum_{j:y_j=y_i}} (D_{ij}^2 + 2m D_{ij})}_{\propto \ent(\widehat{Z}|Y)}  - \underbrace{\frac{2m}{n} \sum_{i=1}^n\sum_{j=1}^n D_{ij}}_{\propto \ent(\widehat{Z})} \quad \propto \; - \mathcal{I}(\widehat{Z}; Y)
\end{equation}

\subsection{Generalizing to other pairwise losses}

\begin{table}
\caption{
Several well-known and/or recent DML losses broken into a \textit{tightness} term and a \textit{contrastive} term. Minimizing the cross-entropy corresponds to an approximate bound optimization of PCE. 
}
\label{table:losses}
\begin{tabular}{lcc}
\textbf{Loss} & \textbf{Tightness part} $\propto\mathcal{H}(\widehat{Z}|Y)$ & \textbf{Contrastive part} $\propto\mathcal{H}(\widehat{Z})$ \\ 
\toprule
Center \cite{center_loss} 
& $ \dfrac{1}{2}\smashoperator{\sum\limits_{i=1}^n} \norm{\z_i - \cc_{y_i}}^2 $
& $-\dfrac{1}{n}\smashoperator{\sum\limits_{i=1}^n} \log p_{iy_i}$
\\
\midrule
Contrast \cite{hadsell2006dimensionality} 
& $\dfrac{1}{n}\smashoperator{\sum\limits_{i=1}^n}\smashoperator[r]{\sum\limits_{j:y_j=y_i}} D_{ij}^2 $ 
& $\dfrac{1}{n}\smashoperator{\sum\limits_{i=1}^n}\smashoperator[r]{\sum\limits_{j:y_j\neq y_i}} [m - D_{ij}]_{+}^2 $ \\
\midrule
SNCA \cite{wu2018improving} 
& $- \dfrac{1}{n}\smashoperator[l]{\sum\limits_{i=1}^n} \log\Bigg[\smashoperator[r]{\sum\limits_{j:y_j=y_i}} \exp{\dfrac{\Dcos_{ij}}{\sigma}}\Bigg]$ 
&$ \dfrac{1}{n}\smashoperator[l]{\sum\limits_{i=1}^n} \log\Bigg[\sum\limits_{k\neq i}\exp{\dfrac{\Dcos_{ik}}{\sigma}}\Bigg]$ \\
\midrule
MS \cite{wang2019multi} 
& $\dfrac{1}{n}\smashoperator[l]{\sum\limits_{i=1}^n} \dfrac{1}{\alpha} \log\Bigg[1 + \smashoperator{\sum\limits_{j:y_j=y_i}} e^{-\alpha(\Dcos_{ij}-m)}\Bigg]$
& $\dfrac{1}{n}\smashoperator[l]{\sum\limits_{i=1}^n} \dfrac{1}{\beta} \log\Bigg[1 + \smashoperator{\sum\limits_{j:y_j \neq y_i}} e^{\beta(\Dcos_{ij} - m)}\Bigg]$ \\
\midrule
\begin{tabular}{l}
     PCE\\
     Prop. \ref{prop:cross_entropy_lower_bound}
\end{tabular}
& $-\dfrac{1}{2\lambda n^2} \smashoperator[l]{\sum\limits_{i=1}^n} \sum_{j:y_j=y_i} \z_i\tran  \z_j$ 
& \begin{tabular}{c}
$\dfrac{1}{n} \sum\limits_{i=1}^n \log\Bigg[\sum\limits_{k=1}^K \exp\Big[\frac{1}{\lambda n} \sum\limits_{j=1}^n p_{jk}\z_i\tran  \z_j\Big]\Bigg]$ \\
$-\dfrac{1}{2 K^2\lambda^2} \smashoperator[l]{\sum\limits_{k=1}^K} \left\| \cc_k^s \right\|^2$\end{tabular} \\
\bottomrule
\end{tabular}
\end{table}

A similar analysis can be carried out on other, more recent metric learning losses. More specifically, they can also be broken down into two parts: a \textit{tightness} part that minimizes intra-class distances to form compact clusters, which is related to the \emph{conditional entropy} $\ent(\widehat{Z}|Y)$, and a second \emph{contrastive} part that prevents trivial solutions by maximizing inter-class distances, which is related to the \emph{entropy} of features $\ent(\widehat{Z})$. Note that, in some pairwise losses, there might be some redundancy between the two terms, \ie, the tightness term also contains some contrastive subterm, and vice-versa. For instance, the cross-entropy loss is used as the contrastive part of the center-loss but, as we show in \autoref{sec:cross_entropy_mi}, the cross-entropy, used alone, already contains both tightness (conditional entropy) and contrastive (entropy) parts. \autoref{table:losses} presents the split for four DML losses. The rest of the section is devoted to exhibiting the close relationships between several pairwise losses and the tightness and contrastive terms (\ie, $T$ and  $C$).

\textbf{Links between losses:}
In this section, we show that the tightness and contrastive parts of the pairwise losses in \autoref{table:losses}, even though different at first sight, can actually be related to one another.\\

\begin{lemma}
\label{prop:tightness_terms}
Let $T_A$ denote the tightness part of the loss from method A. Assuming  that features are $\ell_2$-normalized, and that classes are balanced, the following relations between Center \cite{center_loss}, Contrastive \cite{hadsell2006dimensionality}, SNCA \cite{wu2018improving} and MS \cite{wang2019multi} losses hold:
\begin{align}
    T_\text{SNCA} \cleq T_\text{Center} \ceq T_\text{Contrastive} \cleq T_\text{MS}
\end{align}
Where $\cleq$ stands for lower than, up to a multiplicative and an additive constant, and $\ceq$ stands for equal to, up to a multiplicative and an additive constant.
\end{lemma}

The detailed proof of \autoref{prop:tightness_terms} is deferred to the supplemental material.
As for the contrastive parts, we show in the supplemental material that both $C_{SNCA}$ and $C_{MS}$ are lower bounded by a common contrastive term that is directly related to $\ent(\hat{Z})$. We do not mention the \textit{contrastive} term of center-loss, as it represents the cross-entropy loss, which is exhaustively studied in \autoref{sec:ce}.

%% file: 4_crossentropy.tex
\section{Cross-entropy does it all}\label{sec:ce}

We now completely change gear to focus on the widely used {\em unary} classification loss: cross-entropy. On the surface, the cross-entropy may seem unrelated to metric-learning losses as it does not involve pairwise distances. We show that a close relationship exists between these pairwise losses widely used in deep metric learning and the cross-entropy classification loss. This link can be drawn from two different perspectives, one is based on an explicit optimization insight and the other is based on a discriminative view of the mutual information. First, we explicitly demonstrate that the cross-entropy is an upper bound on a new pairwise loss, which has a structure similar to all the metric-learning losses listed in  \autoref{table:losses}, \ie, it contains a tightness term and a contrastive term. Hence, minimizing the cross-entropy can be seen as an approximate {\em bound-optimization (or Majorize-Minimize)} algorithm for minimizing this pairwise loss. Second, we show that, more generally, minimization of the cross-entropy is actually equivalent to maximization of the mutual information, to which we connected various DML losses. These findings indicate that the cross-entropy represents a proxy for maximizing $\mathcal{I}(\widehat{Z},Y)$, just like pairwise losses, without the need for dealing with the complex sample mining and optimization schemes associated to the latter.

\subsection{The pairwise loss behind unary cross-entropy}\label{sec:cross_entropy_explicit}

\textbf{Bound optimization:}
Given a function $f(\mathcal{W})$ that is either intractable or hard to optimize, bound optimizers are iterative algorithms that instead optimize
auxiliary functions (upper bounds on $f$). These auxiliary functions are usually more tractable than the
original function $f$. Let $t$ be the current iteration index, then $a_t$ is an auxiliary function if:
\begin{align}
\begin{split}
    f(\mathcal{W}) &\leq a_t(\mathcal{W}) \quad,\forall \ \mathcal{W} \\
    f(\mathcal{W}_t) &= a_t(\mathcal{W}_t)
\end{split}
\end{align}
A bound optimizer follows a two-step procedure: first an auxiliary function $a_t$ is computed, then $a_t$ is minimized, such that:
\begin{equation}
    \mathcal{W}_{t+1} = \argmin_{\mathcal{W}} a_t(\mathcal{W})
\end{equation}
This iterative procedure is guaranteed to decrease the original function $f$:
\begin{align}
    f(\mathcal{W}_{t+1}) \leq a_t(\mathcal{W}_{t+1}) \leq a_t(\mathcal{W}_{t}) = f(\mathcal{W}_{t})
\end{align}
Note that bound optimizers are widely used in machine learning. Examples of well-known bound optimizers include the concave-convex procedure (CCCP) \cite{cccp}, expectation maximization (EM) algorithms or submodular-supermodular procedures (SSP) \cite{ssp}. Such optimizers are particularly used in clustering \cite{tang2015kernel} and, more generally, in problems involving latent-variable optimization.
 
\textbf{Pairwise Cross-Entropy:} We now prove that  minimizing cross-entropy can be viewed as an approximate bound optimization of a more complex pairwise loss. 
\begin{proposition}\label{prop:cross_entropy_lower_bound}
Alternately minimizing the cross-entropy loss $\mathcal{L}_{CE}$ with respect to the encoder's parameters $\mathcal{W}$ and the classifier's weights $\w$ can be viewed as an approximate bound-optimization of a Pairwise Cross-Entropy (PCE) loss, which we define as follows:
\begin{equation}
      \mathcal{L}_\text{PCE} = \underbrace{-\frac{1}{2\lambda n^2} \sum_{i=1}^n \smashoperator[r]{\sum_{j:y_j=y_i}} \z_i\tran  \z_j}_{\textsc{tightness part}} + \underbrace{\frac{1}{n} \sum_{i=1}^n \log \sum_{k=1}^K e^{\frac{1}{\lambda n} \sum\limits_{j=1}^n p_{jk}\z_i\tran  \z_j} - \frac{1}{2\lambda } \sum_{k=1}^K \norm{\cc_k^s }^2}_{\textsc{contrastive part}} 
\end{equation}
Where $\cc_k^s=\frac{1}{n}\sum_{i=1}^n p_{ik} \z_i$ represents the soft-mean of class $k$, $p_{ik}$ represents the softmax probability of point $z_i$ belonging to class k, and $\lambda \in \mathbb{R}, \lambda>0$ depends on the encoder $\phi_\mathcal{W}$.
\end{proposition}


The full proof of \autoref{prop:cross_entropy_lower_bound} is provided in the supplemental material. We hereby provide a quick sketch. 
    Considering the usual softmax parametrization for our model's predictions $\widehat{Y}$, the idea is to break the cross-entropy loss in two terms, and artificially add and remove the regularization term $\frac{\lambda}{2}\sum_{k=1}^K \w_k\tran \w_k $:
\begin{equation}\label{eq:ce_split}
    \mathcal{L}_\text{CE} = \underbrace{-\frac{1}{n} \sum_{i=1}^n \w_{y_i}\tran  \z_i + \frac{\lambda}{2}\sum_k \w_k\tran \w_k}_{f_1(\w)} + \underbrace{\frac{1}{n} \sum_{i=1}^n \log\sum_{k=1}^K e^{\w_k\tran  \z_i} - \frac{\lambda}{2}\sum_{k=1}^K \w_k\tran \w_k}_{f_2(\w)}
\end{equation}
By properly choosing $\lambda \in \mathbb{R}$ in Eq. \eqref{eq:ce_split}, both $f_1$ and $f_2$ become convex functions of $\w$. For any class $k$, we then show that the optimal values of $\w_k$ for $f_1$ and $f_2$ are proportional to, respectively, the hard mean $\cc_k=\frac{1}{|\mathcal{Z}_k|}\sum_{i:y_i=k} \z_i$ and the soft mean $\cc_k^s=\frac{1}{n}\sum_{i=1}^n p_{ik} \z_i$ of class $k$. By plugging-in those optimal values, we can lower bound $f_1$ and $f_2$ individually in \autoref{eq:ce_split} and get the result.

\autoref{prop:cross_entropy_lower_bound} casts a new light on the cross-entropy loss by explicitly relating it to a new pairwise loss (PCE), following the intuition that the optimal weights $\w^*$ of the final layer, \ie, the linear classifier, are related to the centroids of each class in the embedded feature space $\mathcal{Z}$. Specifically, finding the optimal classifier's weight $\w^*$ for cross-entropy can be interpreted as building an auxiliary function $a_t(\mathcal{W})=\mathcal{L}_{CE}(\mathcal{W}, \w^*)$ on $\mathcal{L}_{PCE}(\mathcal{W})$. Subsequently minimizing cross-entropy \wrt the encoder's weights $\mathcal{W}$ can be interpreted as the second step of bound optimization on $\mathcal{L}_{PCE}(\mathcal{W})$. 
Similarly to other metric learning losses, PCE contains a $\textit{tightness}$ part that encourages samples from the same classes to align with one another. In echo to \autoref{prop:tightness_terms}, this tightness term, noted $T_\text{PCE}$, is equivalent, up to multiplicative and additive constants, to $T_\text{center}$ and $T_\text{contrast}$, when the features are assumed to be normalized:
\begin{align}
    T_\text{PCE} \ceq T_\text{center} \ceq T_\text{contrast}
\end{align}
 PCE also contains a $\textit{contrastive}$ part, divided into two  terms. The first pushes all samples away from one another, while the second term forces soft means $c_k^s$ far from the origin. Hence, minimizing the cross-entropy can be interpreted as implicitly minimizing a pairwise loss whose structure appears similar to the well-established metric-learning losses in \autoref{table:losses}. \\
 
 \textbf{Simplified Pairwise Cross-Entropy:} While PCE brings interesting theoretical insights, the computation of the parameter $\lambda$ at every iteration requires computating the eigenvalues of a $d \times d$ matrix at every iteration (cf. full proof in supplemental material), which makes the implementation of PCE difficult in practice.
In order to remove the dependence upon $\lambda$, one can plug in the same $\w$ for both $f_1$ and $f_2$ in \autoref{eq:ce_split}. We choose to use $\w_1^{^*}=\underset{\w}{\argmin} f_1(\w) \propto [\cc_1, ..., \cc_K]\tran$. This yields a simplified version of PCE, that we call SPCE:
\newcommand{\lspce}{\mathcal{L}_{SPCE} = \underbrace{-\frac{1}{n^2} \sum_{i=1}^n\sum_{j:y_j=y_i} \z_i\tran \z_j}_{\textsc{tightness}} + \underbrace{\frac{1}{n} \sum_{i=1}^n \log\sum_{k=1}^K \exp \Big(\frac{1}{n} \sum\limits_{j:y_j=k}\z_i\tran \z_j\Big)}_{\textsc{contrastive}}}
\begin{equation}\label{eq:spce}
    \lspce
\end{equation}
SPCE and PCE are similar (the difference is that PCE was derived after plugging in the soft means instead of hard means in $f_2$). Contrary to PCE, however, SPCE is easily computable, and the preliminary experiments we provide in the supplementary material indicate that CE and SPCE exhibit similar behaviors at training time. 
Interestingly, our derived SPCE loss has a form similar to contrastive learning losses in unsupervised representation learning \cite{oord2018representation, tschannen2019mutual, chen2020simple}.

\subsection{A discriminative view of mutual information}\label{sec:cross_entropy_mi}

\begin{lemma}\label{prop:cross_ent_and_mi}
Minimizing the conditional cross-entropy loss, denoted by  $\ent(Y;\widehat{Y}|\widehat{Z})$, is equivalent to maximizing the mutual information $\mathcal{I}(\widehat{Z}; Y)$.
\end{lemma}
The proof of \autoref{prop:cross_ent_and_mi} is provided in the supplementary material. Such result is compelling. Using the discriminative view of mutual information allows to show that minimizing cross-entropy loss is equivalent to maximizing the mutual information $\mathcal{I}(\widehat{Z}; Y)$. This information theoretic argument reinforces our conclusion from \autoref{prop:cross_entropy_lower_bound} that cross-entropy and the previously described metric learning losses are essentially doing the same job.

\subsection{Then why would cross-entropy work better?}

We showed that cross-entropy essentially optimizes the same underlying mutual information $\mathcal{I}(\widehat{Z};Y)$ as other DML losses. This fact alone is not enough to explain why the cross-entropy is able to consistently achieve better results than DML losses as shown in \autoref{sec:exps}. We argue that the difference is in the optimization process. On the one hand, pairwise losses require careful sample mining and weighting strategies to obtain the most informative pairs, especially when considering mini-batches, in order to achieve convergence in a reasonable amount of time, using a reasonable amount of memory. On the other hand, optimizing cross-entropy is substantially easier as it only implies minimization of unary terms. 
Essentially, cross-entropy does it all without dealing with the difficulties of pairwise terms.
Not only it makes optimization easier, but also it simplifies the implementation, thus increasing its potential applicability in real-world problems.

%% file: 5_0_experiments.tex
\section{Experiments}
\label{sec:exps}

\subsection{Metric}

Most methods, especially recent ones, use the cosine distance to compute the recall for the evaluation. They include $\ell_2$ normalization of the features in the model \cite{oh2017deep, movshovitz2017no, wang2017deep, opitz2017bier, ge2018deep, yuan2017hard, xuan2020improved, zhai2018classification, wang2019multi, sanakoyeu2019divide, xuan2018deep}, which makes cosine and Euclidean distances equivalent. Computing cosine similarity is also more memory efficient and typically leads to better results \cite{schroff2015facenet}. For these reasons, the Euclidean distance on non normalized features has rarely been used for both training and evaluation.
In our experiments, $\ell_2$-normalization of the features during training actually hindered the final performance, which might be explained by the fact that we add a classification layer on top of the feature extractor. Thus, we did not $\ell_2$-normalize the features during training and reported the recall with both Euclidean and cosine distances.

\subsection{Datasets}

\setlength{\tabcolsep}{5pt}
\begin{table}[t]
\centering
\caption{Summary of the datasets used for evaluation in metric learning.}
\label{table:datasets}
\begin{tabularx}{\textwidth}{lccc}
Name & Objects & Categories & Images \\ 
\toprule
Caltech-UCSD Birds-200-2011 (CUB)\cite{WahCUB_200_2011} & Birds & 200 & 11\,788 \\
Cars Dataset \cite{krause20133d} & Cars & 196 & 16\,185 \\
Stanford Online Products (SOP) \cite{song2016deep} & House furniture & 22\,634 & 120\,053 \\
In-shop Clothes Retrieval \cite{liu2016deepfashion} & Clothes & 7\,982 & 52\,712 \\
\bottomrule
\end{tabularx}
\end{table}
Four datasets are commonly used in metric learning to evaluate the performances. These datasets are summarized in \autoref{table:datasets}. CUB \cite{WahCUB_200_2011}, Cars \cite{krause20133d} and SOP \cite{song2016deep} datasets are divided into train and evaluation splits. For the evaluation, the recall is computed between each sample of the evaluation set and the rest of the set. In-Shop \cite{liu2016deepfashion} is divided into a query and a gallery set. The recall is computed between each sample of the query set and the whole gallery set.

\subsection{Training specifics}

\textbf{Model architecture and pre-training:}
In the metric learning literature, several architectures have been used, which historically correspond to the state-of-the-art image classification architectures on ImageNet \cite{deng2009imagenet}, with an additional constraint on model size (\ie, the ability to train on one or two GPUs in a reasonable time). These include GoogLeNet \cite{szegedy2015going} as in \cite{kim2018attention}, BatchNorm-Inception \cite{szegedy2016rethinking} as in \cite{wang2019multi} and ResNet-50 \cite{he2016identity} as in \cite{xuan2020improved}. They have large differences in classification performances on ImageNet, but the impact on performances over DML benchmarks has rarely been studied in controlled experiments. As this is not the focus of our paper, we use ResNet-50 for our experiments. We concede that one may obtain better performances by modifying the architecture (\eg, reducing model stride and performing multi-level fusion of features). Here, we limit our comparison to standard architectures. Our implementation uses the PyTorch \cite{NIPS2019_9015} library, and initializes the ResNet-50 model with  weights pre-trained on ImageNet.


\input{5_1_all_tables}
\textbf{Sampling:} To the best of our knowledge, all DML papers -- including \cite{zhai2018classification} -- use a form of pairwise sampling to ensure that, during training, each mini-batch contains a fixed number of classes and samples per class (\eg mini-batch size of 75 with 3 classes and 25 samples per class in \cite{zhai2018classification}). Deviating from that, we use the common random sampling among all samples (as in most classification training schemes) and set the mini-batch size to 128 in all experiments (contrary to \cite{wang2019multi} in which the authors use a mini-batch size of 80 for CUB, 1\,000 for SOP and did not report for Cars and In-Shop).

\textbf{Data Augmentation:} As is common in training deep learning models, data augmentation improves the final performances of the methods. 
For CUB, the images are first resized so that their smallest side has a length of 256 (\ie, keeping the aspect ratio) while for Cars, SOP and In-Shop, the images are resized to $256\times 256$. Then a patch is extracted at a random location and size, and resized to $224\times 224$. For CUB and Cars, we found that random jittering of the brightness, contrast and saturation slightly improves the results. All of the implementation details can be found in the publicly available code.



\textbf{Cross-entropy:} The focus of our experiments is to show that, with careful tuning, it is possible to obtain similar or better performance than most recent DML methods, while using only the cross-entropy loss. To train with the cross-entropy loss, we add a linear classification layer (with bias) on top of the feature extraction -- similar to many classification models -- which produces logits for all the classes present in the training set. Both the weights and biases of this classification layer are initialized to $\bm{0}$. We also add dropout with a probability of $0.5$ before this classification layer. To further reduce overfitting, we use label smoothing for the target probabilities of the cross-entropy. We set the probability of the true class to $1-\epsilon$ and the probabilities of the other classes to $\frac{\epsilon}{K-1}$ with $\epsilon=0.1$ in all our experiments.

\textbf{Optimizer:} In most DML papers, the hyper-parameters of the optimizer are the same for Cars, SOP and In-Shop whereas, for CUB, the methods typically use a smaller learning rate. In our experiments, we found that the best results were obtained by tuning the learning rate on a per dataset basis. In all experiments, the models are trained with SGD with Nesterov acceleration and a weight decay of $0.0005$, which is applied to convolution and fully-connected layers' weights (but not to biases) as in \cite{jia2018highly}. For CUB and Cars, the learning rate is set to 0.02 and 0.05 respectively, with 0 momentum. For both SOP and In-Shop, the learning rate is set to 0.003 with a momentum of 0.99.

\textbf{Batch normalization:} Following \cite{wang2019multi}, we freeze all the batch normalization layers in the feature extractor. For Cars, SOP and In-Shop, we found that adding batch normalization -- without scaling and bias -- on top of the feature extractor improves our final performance and reduces the gap between $\ell_2$ and cosine distances when computing the recall. On CUB, however, we obtained the best recall without this batch normalization.


\subsection{Results}

Results for the experiments are reported in \autoref{table:results}. We also report the architecture used in the experiments as well as the distance used in the evaluation to compute the recall. \emph{$\ell_2$} refers to the Euclidean distance on non normalized features while \emph{cos} refers to either the cosine distance or the Euclidean distance on $\ell_2$-normalized features, both of which are equivalent.

On all datasets, we report state-of-the-art results except on Cars, where the only method achieving similar recall uses cross-entropy for training. 
We also notice that, contrary to common beliefs, using Euclidean distance can actually be competitive as it also achieves near state-of-the-art results on all four datasets. These results clearly highlight the potential of cross-entropy for metric learning, and confirm that this loss can achieve the same objective as pairwise losses.

%% file: 5_1_all_tables.tex
\begin{table}
\centering
\caption{Performance on CUB200, Cars-196, SOP and In-Shop datasets. $d$ refers to the distance used to compute the recall when evaluating.}
\label{table:results}
\resizebox{\textwidth}{!}{
\begin{tabular}{clccc}
& Method & $d$ & Architecture & Recall at \\
\toprule
\multirow{13}{*}{\begin{turn}{90}Caltech-UCSD Birds-200-2011\end{turn}} & & & & 
\multirow{13}{*}{\begin{tabular}{cccccc} 1 & 2 & 4 & 8 & 16 & 32\\
47.2 & 58.9 & 70.2 & 80.2 & 89.3 & 93.2 \\
49.2 & 61.9 & 67.9 & 81.9 & - & - \\
57.1 & 68.8 & 78.7 & 86.5 & 92.5 & 95.5 \\
60.6 & 71.5 & 79.8 & 87.4 & -- & -- \\
60.7 & 72.4 & 81.9 & 89.2 & 93.7 & 96.8 \\
63.9 & 75.0 & 83.1 & 89.7 & -- & -- \\
64.9 & 75.3 & 83.5 & -- & -- & --\\
65.3 & 76.7 & 85.4 & 91.8 & -- & -- \\
65.7 & 77.0 & 86.6 & 91.2 & 95.0 & 97.3 \\
65.9 & 76.6 & 84.4 & 90.6 & -- & --\\
67.6 & 78.1 & 85.6 & 91.1 & 94.7 & 97.2 \\
69.2 & 79.2 & 86.9 & 91.6 & 95.0 & 97.3 \\
\end{tabular}}\\
\cline{5-5}
& Lifted Structure \cite{song2016deep} & $\ell_2$ & GoogLeNet & \\
& Proxy-NCA \cite{movshovitz2017no} & cos & BN-Inception & \\
& HTL \cite{ge2018deep} & cos & GoogLeNet & \\
& ABE \cite{kim2018attention} & cos & GoogLeNet & \\
& HDC \cite{yuan2017hard} & cos & GoogLeNet & \\
& DREML \cite{xuan2018deep} & cos & ResNet-18 & \\
& EPSHN \cite{xuan2020improved} & cos & ResNet-50 & \\
& NormSoftmax \cite{zhai2018classification} & cos & ResNet-50 & \\
& Multi-Similarity \cite{wang2019multi} & cos & BN-Inception& \\
& D\&C \cite{sanakoyeu2019divide} & cos & ResNet-50 & \\
\cline{2-5}
& \multirow{2}{*}{Cross-Entropy} & $\ell_2$ & \multirow{2}{*}{ResNet-50} & \\
& & cos & & \\
\midrule[0.75pt]
\multirow{13}{*}{\begin{turn}{90}Stanford Cars\end{turn}} & & & & 
\multirow{13}{*}{\begin{tabular}{cccccc} 1 & 2 & 4 & 8 & 16 & 32\\
49.0 & 60.3 & 72.1 & 81.5 & 89.2 & 92.8 \\
73.2 & 82.4 & 86.4 & 88.7 & -- & -- \\
81.4 & 88.0 & 92.7 & 95.7 & 97.4 & 99.0 \\
82.7 & 89.3 & 93.0 & -- & -- & --\\
83.8 & 89.8 & 93.6 & 96.2 & 97.8 & 98.9 \\
84.1 & 90.4 & 94.0 & 96.5 & 98.0 & 98.9 \\
84.6 & 90.7 & 94.1 & 96.5 & -- & -- \\
85.2 & 90.5 & 94.0 & 96.1 & -- & -- \\
86.0 & 91.7 & 95.0 & 97.2 & -- & -- \\
89.3 & 94.1 & 96.4 & 98.0 & -- & -- \\
89.1 & 93.7 & 96.5 & 98.1 & 99.0 & 99.4 \\
89.3 & 93.9 & 96.6 & 98.4 & 99.3 & 99.7 \\
\end{tabular}}\\
\cline{5-5}
& Lifted Structure \cite{song2016deep} & $\ell_2$ & GoogLeNet & \\
& Proxy-NCA \cite{movshovitz2017no} & cos & BN-Inception \\
& HTL \cite{yuan2017hard}  & cos & GoogLeNet & \\
& EPSHN \cite{xuan2020improved} & cos & ResNet-50 & \\
& HDC \cite{yuan2017hard} & cos & GoogLeNet & \\
& Multi-Similarity \cite{wang2019multi} & cos & BN-Inception & \\
& D\&C \cite{sanakoyeu2019divide} & cos & ResNet-50 & \\
& ABE \cite{kim2018attention} & cos & GoogLeNet & \\
& DREML \cite{xuan2018deep} & cos & ResNet-18 & \\
& NormSoftmax \cite{zhai2018classification} & cos & ResNet-50 &  \\
\cline{2-5}
& \multirow{2}{*}{Cross-Entropy} & $\ell_2$ & \multirow{2}{*}{ResNet-50} & \\
& & cos & & \\
\midrule[0.75pt]
\multirow{11}{*}{\begin{turn}{90}Stanford Online Product\end{turn}} & & & & 
\multirow{11}{*}{\setlength{\tabcolsep}{13.5pt}\begin{tabular}{cccc} 1 & 10 & 100 & 1000\\
62.1 & 79.8 & 91.3 & 97.4 \\
70.1 & 84.9 & 93.2 & 97.8 \\
74.8 & 88.3 & 94.8 & 98.4 \\
75.9 & 88.4 & 94.9 & 98.1 \\
76.3 & 88.4 & 94.8 & 98.2 \\
78.2 & 90.5 & 96.0 & 98.7 \\
78.3 & 90.7 & 96.3 & -- \\
79.5 & 91.5 & 96.7 & -- \\
80.8 & 91.2 & 95.7 & 98.1 \\
81.1 & 91.7 & 96.3 & 98.8 \\
\end{tabular}}
\setlength{\tabcolsep}{1pt}\\
\cline{5-5}
& Lifted Structure \cite{song2016deep} & $\ell_2$ & GoogLeNet & \\
& HDC \cite{yuan2017hard} & cos & GoogLeNet & \\
& HTL \cite{ge2018deep} & cos & GoogLeNet & \\
& D\&C \cite{sanakoyeu2019divide} & cos & ResNet-50 & \\
& ABE \cite{kim2018attention} & cos & GoogLeNet & \\
& Multi-Similarity \cite{wang2019multi} & cos & BN-Inception &\\
& EPSHN \cite{xuan2020improved} & cos & ResNet-50 & \\
& NormSoftmax \cite{zhai2018classification} & cos & ResNet-50 & \\
\cline{2-5}
& \multirow{2}{*}{Cross-Entropy} & $\ell_2$ & \multirow{2}{*}{ResNet-50} & \\
& & cos & & \\
\midrule[0.75pt]
\multirow{11}{*}{\begin{turn}{90}In-Shop Clothes Retrieval\end{turn}} & & & & 
\multirow{11}{*}{\begin{tabular}{cccccc} 1 & 10 & 20 & 30 & 40 & 50\\
62.1 & 84.9 & 89.0 & 91.2 & 92.3 & 93.1 \\
78.4 & 93.7 & 95.8 & 96.7 & -- & -- \\
80.9 & 94.3 & 95.8 & 97.2 & 97.4 & 97.8 \\
85.7 & 95.5 & 96.9 & 97.5 & -- & 98.0 \\
87.3 & 96.7 & 97.9 & 98.2 & 98.5 & 98.7 \\
87.8 & 95.7 & 96.8 & -- & -- & -- \\
89.4 & 97.8 & 98.7 & 99.0 & -- & -- \\
89.7 & 97.9 & 98.5 & 98.8 & 99.1 & 99.2 \\
90.6 & 97.8 & 98.5 & 98.8 & 98.9 & 99.0 \\
90.6 & 98.0 & 98.6 & 98.9 & 99.1 & 99.2 \\
\end{tabular}}\\
\cline{5-5}
& HDC \cite{yuan2017hard} & cos & GoogLeNet & \\
& DREML \cite{xuan2018deep} & cos & ResNet-18 & \\
& HTL \cite{ge2018deep} & cos & GoogLeNet & \\
& D\&C \cite{sanakoyeu2019divide} & cos & ResNet-50 & \\
& ABE \cite{kim2018attention} & cos & GoogLeNet & \\
& EPSHN \cite{xuan2020improved} & cos & ResNet-50 & \\
& NormSoftmax \cite{zhai2018classification} & cos & ResNet-50 & \\
& Multi-Similarity \cite{wang2019multi} & cos & BN-Inception & \\
\cline{2-5}
& \multirow{2}{*}{Cross-Entropy} & $\ell_2$ & \multirow{2}{*}{ResNet-50} & \\
& & cos & & \\
\bottomrule
\end{tabular}
}
\end{table}

%% file: 6_conclusion.tex
\section{Conclusion}

Throughout this paper, we revealed non-obvious relations between the cross-entropy loss, widely adopted in classification tasks, and pairwise losses commonly used in DML. These relations were drawn under two different perspectives. First, cross-entropy minimization was shown equivalent to an approximate bound-optimization of a pairwise loss, introduced as Pairwise Cross-Entropy (PCE), which appears similar in structure to already existing DML losses. Second, adopting a more general information theoretic view of DML, we showed that both pairwise losses and cross-entropy were, in essence, maximizing a common mutual information $\mathcal{I}(\hat{Z}, Y)$ between the embedded features and the labels. This connection becomes particularly apparent when writing mutual information in both its \textit{generative} and \textit{discriminative} views. Hence, we argue that most of the differences in performance observed in previous works come from the optimization process during training. Cross-entropy contains only unary terms, while traditional DML losses are based on pairwise-term optimization, which requires substantially more tuning (\eg mini-batch size, sampling strategy, pair weighting).
While we acknowledge that some losses have better properties than others regarding optimization, we empirically showed that the cross-entropy loss was also able to achieve state-of-the-art results when fairly tuned, highlighting the fact that most improvements have come from enhanced training schemes (\eg data augmentation, learning rate policies, batch normalization freeze) rather than the intrinsic properties of pairwise losses.
We strongly advocate that cross-entropy should be carefully tuned to be compared against as a baseline in future works. 

%% file: 7_appendix.tex
\appendix
\section{Proofs}

    \subsection{\autoref{prop:tightness_terms}}\label{appendix:proof_tightness_terms}
        \begin{proof}
            Throughout the following proofs, we will use the fact that classes are assumed to be balanced in order to consider $\mathcal{Z}_k$, for any class $k$, as a constant $|\mathcal{Z}_k|=\frac{n}{K}$. We will also use the feature normalization assumption to connect cosine and Euclidean distances. On the unit-hypersphere, we will use that: $D_{i,j}^{\text{cos}} = 1 - \frac{\left\| \z_i - \z_j \right\|^2}{2}$.\\
            
            \subsubsection{Tightness terms:}
            Let us start by linking center loss to contrastive loss. For any specific class $k$, let  $\cc_k = \frac{1}{|\mathcal{Z}_k|} \sum\limits_{z \in \mathcal{Z}_k} z$ denotes the hard mean. We can write:
            \begin{align*}
                \sum_{\z_i \in \mathcal{Z}_k} \left\| \z_i - \cc_k \right\|^2
                &= \sum_{\z_i \in \mathcal{Z}_k} [\left\| \z_i \right\|^2 - 2 \z_i\tran \cc_k] +  |\mathcal{Z}_k| \left\| \cc_k \right\|^2 \\
                &= \sum_{\z_i \in \mathcal{Z}_k} \left\| \z_i \right\|^2 - 2\frac{1}{|\mathcal{Z}_k|} \sum_{\z_i \in \mathcal{Z}_k} \sum_{\z_j \in \mathcal{Z}_k} \z_i\tran \z_j + \frac{1}{|\mathcal{Z}_k|}\sum_{\z_i \in \mathcal{Z}_k}\sum_{\z_j \in \mathcal{Z}_k} \z_i\tran \z_j \\
                &= \sum_{\z_i \in \mathcal{Z}_k} \left\| \z_i \right\|^2 -  \frac{1}{|\mathcal{Z}_k|}\sum_{\z_i \in \mathcal{Z}_k} \sum_{\z_j \in \mathcal{Z}_k} \z_i\tran \z_j \\
                &= \frac{1}{2}[\sum_{\z_i \in \mathcal{Z}_k} \left\| \z_i \right\|^2 + \sum_{\z_j \in \mathcal{Z}_k} \left\| \z_j \right\|^2] - \frac{1}{|\mathcal{Z}_k|}\sum_{\z_i \in \mathcal{Z}_k}\sum_{\z_j \in \mathcal{Z}_k} \z_i\tran \z_j \\
                &= \frac{1}{2|\mathcal{Z}_k|}[\sum_{\z_i \in \mathcal{Z}_k}\sum_{\z_j \in \mathcal{Z}_k} \left\| \z_i \right\|^2 + \sum_{\z_i \in \mathcal{Z}_k}\sum_{\z_j \in \mathcal{Z}_k} \left\| \z_j \right\|^2] \\ &\ \ - \frac{1}{2|\mathcal{Z}_k|}\sum_{\z_i \in \mathcal{Z}_k} \sum_{\z_j \in \mathcal{Z}_k} 2z_i\tran \z_j\\
                &=\frac{1}{2|\mathcal{Z}_k|} \sum_{\z_i, \z_j\in \mathcal{Z}_k} \left\| \z_i \right\|^2 -2 \z_i\tran \z_j + \left\| \z_j \right\|^2 \\
                &= \frac{1}{2|\mathcal{Z}_k|} \sum_{\z_i, \z_j\in \mathcal{Z}_k} \left\| \z_i - \z_j \right\|^2 \\
                & \ceq \sum_{\z_i, \z_j\in \mathcal{Z}_k} \left\| \z_i - \z_j \right\|^2 \\
            \end{align*}
            Summing over all classes $k$, we get the desired equivalence. Note that, in the context of K-means clustering, where the setting is different\footnote{In clustering, the optimization is performed over assignment variables, as opposed to DML, where assignments are already known and optimization is carried out over the embedding.}, a technically similar result could be established \cite{tang2015kernel}, linking K-means to pairwise graph clusteirng objectives. \\
            
            \noindent
            Now we link contrastive loss to SNCA loss. For any class $k$, we can write:
            \begin{align*}
                - \sum_{\z_i \in \mathcal{Z}_k} \log \sum_{\z_j \in \mathcal{Z}_k \setminus \{i\}} e^{\frac{D_{i,j}^{\text{cos}}}{\sigma}}  
                & \ceq  -\sum_{\z_i \in \mathcal{Z}_k} \log\left(\frac{1}{|\mathcal{Z}_k|-1}\sum_{\z_j \in \mathcal{Z}_k \setminus \{i\}} e^{\frac{D_{i,j}^{\text{cos}}}{\sigma}}\right) \\
                & \leq 
                -\sum_{\z_i \in \mathcal{Z}_k} \sum_{\z_j \in \mathcal{Z}_k \setminus \{i\}} \frac{D_{i,j}^{\text{cos}}}{(|\mathcal{Z}_k|-1)\sigma} \\
                & \ceq \sum_{\z_i \in \mathcal{Z}_k} \sum_{\z_j \in \mathcal{Z}_k \setminus \{i\}} \frac{\left\| \z_i - \z_j \right\|^2}{2\sigma(|\mathcal{Z}_k|-1)} \\
                & \ceq \sum_{\z_i \in \mathcal{Z}_k} \sum_{\z_j \in \mathcal{Z}_k \setminus \{i\}} {\left\| \z_i - \z_j \right\|^2}
            \end{align*}
            where we used the convexity of $x \rightarrow{-\log(x)}$ and Jenson's inequality. The proof can be finished by summing over all classes $k$. \\
            
            Finally, we link MS loss \cite{wang2019multi} to contrastive loss:
            \begin{align*}
                \smashoperator{\sum_{\z_i \in \mathcal{Z}_k}} \frac{1}{\alpha} \log\left (1 + \smashoperator{\sum_{\z_j \in \mathcal{Z}_{k} \setminus \{i\}}} e^{-\alpha (D_{i,j}^{\text{cos}} - 1)}\right)
                &= \smashoperator[r]{\sum_{\z_i \in \mathcal{Z}_k}} \frac{1}{\alpha} \log \smashoperator{\sum_{\z_j \in \mathcal{Z}_k}} e^{-\alpha (D_{i,j}^{\text{cos}} - 1)} \\
               & \ceq \smashoperator[r]{\sum_{\z_i \in \mathcal{Z}_k}} \frac{1}{\alpha} \log \left (\frac{1}{|\mathcal{Z}_k|}\smashoperator[r]{\sum_{\z_j \in \mathcal{Z}_k}}e^{-\alpha (D_{i,j}^{\text{cos}} - 1)} \right ) 
                \\
                &\geq 
                 \frac{1}{|\mathcal{Z}_k|} \smashoperator[r]{\sum_{\z_i,z_j \in \mathcal{Z}_k}} -(D_{i,j}^{\text{cos}} - 1) \\
                 & \ceq \smashoperator[r]{\sum_{\z_i,\z_j \in \mathcal{Z}_k}} \left\| \z_i - \z_j \right\|^2,
            \end{align*}
            where we used the concavity of $x \rightarrow{\log(x)}$ and  Jenson's inequality.
            
            \subsubsection{Contrastive terms:} In this part, we first show that the contrastive terms $C_{SNCA}$ and $C_{MS}$ represent upper bounds on $C=-\frac{1}{n}\sum_{i=1}^n \sum_{j:y_j\neq y_i} D_{ij}^2$:
            \begin{align*}
                C_{MS} = \frac{1}{\beta n}\sum_{i=1}^n \log \left (1+\sum_{j:y_j \neq y_i}e^{\beta(\Dcos_{ij}-1)} \right ) &\geq \frac{1}{\beta n}\sum_{i=1}^n \log \left (\sum_{j:y_j \neq y_i}e^{\beta(\Dcos_{ij}-1)} \right ) \\
                &\cgeq \frac{1}{\beta n}\sum_{i=1}^n \sum_{j:y_j \neq y_i}\beta(\Dcos_{ij}-1) \\
                &\ceq -\frac{1}{n}\sum_{i=1}^n \sum_{j:y_j \neq y_i} D_{ij}^2 \\
                &= C
            \end{align*}
            where, again, we used Jenson's inequality in the second line above. The link between SNCA and contrastive loss can be established quite similarly:
            \begin{align}
                C_{SNCA} = \frac{1}{n}\sum_{i=1}^n \log \left (\sum_{j \neq i}e^{\frac{\Dcos_{ij}}{\sigma}} \right ) &= \frac{1}{n}\sum_{i=1}^n \log \left (\sum_{j \neq i:y_i=y_j}e^{\frac{\Dcos_{ij}}{\sigma}} + \sum_{j:y_j\neq y_i}e^{\frac{\Dcos_{ij}}{\sigma}} \right ) \\
                &\geq \frac{1}{n}\sum_{i=1}^n \log \left (\sum_{j:y_j\neq y_i}e^{\frac{\Dcos_{ij}}{\sigma}} \right ) \\
                &\cgeq \frac{1}{n}\sum_{i=1}^n \sum_{j:y_j\neq y_i}\frac{\Dcos_{ij}}{\sigma} \\
                &\ceq -\frac{1}{n}\sum_{i=1}^n \sum_{j:y_j\neq y_i} D_{ij}^2 \\
                &= C
            \end{align}
            
            Now, similarly to the reasoning carried out in \autoref{subsec:contrastive_example}, we can write:
            \begin{align*}
                C=-\frac{1}{n}\sum_{i=1}^n \sum_{j:y_j\neq y_i} D_{ij}^2 = -\underbrace{\frac{1}{n}\sum_{i=1}^n \sum_{j=1}^n D_{ij}^2}_{\text{contrast } \propto \ \ent(\hat{Z})} + \underbrace{\frac{1}{n}\sum_{i=1}^n \sum_{j:y_j = y_i} D_{ij}^2}_{\text{tightness subterm }\ \propto \ \ent(\hat{Z}|Y)}
            \end{align*}
            Where the redundant tightness term is very similar to the tightness term in contrastive loss $T_{contrast}$ treated in details in \autoref{subsec:contrastive_example}. As for the truly contrastive part of $C$, it can also be related to the differential entropy estimator used in \cite{wang2011information}:
            \begin{equation}
                \widehat{\ent}(\widehat{Z}) = \frac{d}{n(n-1)} \sum_{i=1}^n\sum_{j=1}^n \log D_{ij}^2 \ \ceq \ \frac{1}{n} \sum_{i=1}^n\sum_{j=1}^n \log D_{ij}^2
            \end{equation}
            In summary, we just proved that the contrastive parts of MS and SNCA losses are upper bounds on the contrastive term $C$. The latter term is composed of a proxy for the entropy of features $\ent(\hat{Z})$, as well 
            as a tightness sub-term.
        \end{proof}

    \subsection{\autoref{prop:cross_entropy_lower_bound}}\label{appendix:proof_cross_entropy_lower_bound}
        \begin{proof}
        First, let us show that $\mathcal{L}_{CE} \geq \mathcal{L}_{PCE}$. 
        Consider the usual softmax parametrization of point $i$ belonging to class $k$: $p_{ik}=(f_{\w}(z_i))_k=\frac{\exp{\w_k\tran \z_i}}{\sum_j \exp{\w_j\tran \z_i}}$, where $z=\phi_\mathcal{W}(x)$. We can explicitly write the cross-entropy loss:
        \begin{align}
            \mathcal{L}_{CE} &= -\frac{1}{n} \sum_{i=1}^n \log f_{\w}(z_i) \nonumber \\ 
            &= \underbrace{-\frac{1}{n} \sum_{i=1}^n \w_{y_i}\tran \z_i + \frac{\lambda}{2}\sum_{k=1}^K \w_k\tran\w_k}_{f_1(\w)} + \underbrace{\frac{1}{n} \sum_{i=1}^n \log\sum_{j=1}^K e^{\w_j\tran \z_i} - \frac{\lambda}{2}\sum_{k=1}^K \w_k\tran\w_k}_{f_2(\w)}. \label{eq:cross_entropy_explicit}
        \end{align}
        Where we introduced $\lambda \in \mathbb{R}$. How to specifically set $\lambda$ will soon become clear. Let us now write the gradients of $f_1$ and $f_2$ in \autoref{eq:cross_entropy_explicit} with respect to $\w_k$:
        \begin{align}
            \frac{\partial f_1}{\partial \w_k} &= - \frac{1}{n} \smashoperator[r]{\sum_{i:y_i=k}} \z_i + \lambda \w_k \label{eq:gradients_1} \\
            \frac{\partial f_2}{\partial \w_k} &= \frac{1}{n} \sum_{i} \underbrace{\frac{\exp(\w_k\tran \z_i)}{\sum_{j=1}^K \exp(\w_j\tran \z_i)}}_{p_{ik}} \z_i - \lambda \w_k \label{eq:gradients_2}
        \end{align}
        Notice that $f_1$ is a convex function of $\w$, regardless of $\lambda$. As for $f_2$, we set $\lambda$ such that $f_2$ becomes a convex function of $\w$. Specifically, by setting:
        \begin{align}
            \lambda=\min\limits_{k, l} \sigma_l(A_k)
        \end{align}
        where $A_k=\frac{1}{n}\sum_{i=1}^n(p_{ik}-p_{ik}^2)\z_i\z_i\tran$ and $\sigma_l(A)$ represents the $l^{th}$ eigenvalue of $A$, we make sure that the hessian of $f_2$ is semi-definite positive. Therefore, we can look for the minima of $f_1$ and $f_2$.\\
        
        Setting gradients in \autoref{eq:gradients_1} and \autoref{eq:gradients_2} to 0, we obtain that for all $k \in [1, K]$, the optimal $\w_k$ for $f_1$ is, up to a multiplicative constant, the hard mean of features from class $k$: $\w_k^{f_{1}*}=\frac{1}{\lambda n}\underset{i:y_i=k}{\sum}\z_i \propto \cc_k$, while the optimal $\w_k$ for $f_2$ is, up to a multiplicative constant, the soft mean of features: $\w_k^{f_{2}*}=\frac{1}{\lambda n}\sum_{i=1}^n\ p_{ik} \z_i = \cc_k^s/\lambda$. Therefore, we can 
        write:
        \begin{align}
            f_1(\w) \geq f_1(\w^{f_{1}*}) & = - \frac{1}{\lambda n^2} \sum_{i=1}^n \smashoperator[r]{\sum_{j:y_j=y_i}}\z_i\tran \z_j +  \frac{\lambda}{2 \lambda^2} \sum_{i=1}^n \smashoperator[r]{\sum_{j:y_j=y_i}}\z_i\tran \z_j \\
            & = - \frac{1}{2\lambda n^2} \sum_{i=1}^n \smashoperator[r]{\sum_{j:y_j=y_i}}\z_i\tran \z_j
        \end{align}
        And 
        \begin{align}
            f_2(\w) &\geq f_2(\w^{f_{2}*}) \\
            &= \frac{1}{n} \sum_{i=1}^n \log \sum_{k=1}^K \exp\left(\frac{1}{\lambda n} \sum_{j=1}^n p_{jk}\z_i\tran \z_j\right ) - \frac{1}{2 \lambda}\sum_{k=1}^K \left\| \cc_k^s \right\|^2
        \end{align}
        Putting it all together, we can obtain the desired result:
        \begin{align}
            \mathcal{L}_{CE} &\geq - \frac{1}{2\lambda n^2} \sum_{i=1}^n \smashoperator[r]{\sum_{j:y_j=y_i}}\z_i\tran \z_j + \frac{1}{n} \sum_{i=1}^n \log \sum_{k=1}^K e^{\frac{1}{\lambda n} \underset{j}{\sum} p_{jk}\z_i\tran \z_j} - \frac{1}{ 2 \lambda}\sum_{k=1}^K \norm{\cc_k^s}^2 \\
            &= \mathcal{L}_{PCE}
        \end{align}
        where $\cc_k^s=\frac{1}{n} \sum_{i=1}^{n} p_{ik} \z_i$ represents the soft mean of class k.  \\
        
        Let us now justify that minimizing cross-entropy can be seen as an approximate bound optimization on $\mathcal{L}_{PCE}$. At every iteration $t$ of the training, cross-entropy represents an upper bound on Pairwise Cross-entropy.
        
        \begin{align}
            \mathcal{L}_{CE}(\mathcal{W}(t), \w(t)) \geq \mathcal{L}_{PCE}(\mathcal{W}(t), \w(t))
        \end{align}
        When optimizing w.r.t $\theta$, the bound almost becomes tight. The approximation comes from the fact that $\w_k^{f_{1}*}$ and $\w_k^{f_{2}*}$ are quite dissimilar in early training, but become very similar as training progresses and the model's softmax probabilities align with the labels. Therefore, using the notation:
        \begin{align}
            \w(t+1) = \underset{\w}{\min} \ \mathcal{L}_{CE}(\mathcal{W}(t), \w)  
        \end{align}
        We can write:
        \begin{align}
            \mathcal{L}_{CE}(\mathcal{W}(t), \w(t+1)) \approx \mathcal{L}_{PCE}(\mathcal{W}(t), \w(t+1))
        \end{align}
        Then, minimizing $\mathcal{L}_{CE}$ and $\mathcal{L}_{PCE}$ w.r.t $\mathcal{W}$ becomes approximately equivalent.
        \end{proof}
    
    \subsection{\autoref{prop:cross_ent_and_mi}}\label{appendix:cross_ent_and_mi}
        \begin{proof}
            Using the discriminative view of MI, we can write:
            \begin{equation}
                \mathcal{I}(\widehat{Z}; Y) = \ent(Y) - \ent(Y|\widehat{Z})
            \end{equation}
            The entropy of labels $\ent(Y)$ is a constant and, therefore, can be ignored. From this view of MI, maximization of $\mathcal{I}(\widehat{Z}; Y)$ can only be achieved through a minimization of $\ent(Y|\widehat{Z})$, which depends on our embeddings $\widehat{Z}=\phi_\mathcal{W}(X)$. We can relate this term to our cross-entropy loss using the following relation:
            \begin{align}\label{eq:penalty}
                \ent(Y;\widehat{Y}|\widehat{Z}) = \ent(Y|\widehat{Z}) + \mathcal{D}_{KL}(Y\|\widehat{Y}|\widehat{Z})
            \end{align}
            Therefore, while minimizing cross-entropy, we are implicitly both minimizing $\ent(Y|\widehat{Z})$ as well as $\mathcal{D}_{KL}(Y\|\widehat{Y}|\widehat{Z})$.   In fact, following \autoref{eq:penalty}, optimization could naturally be decoupled in 2 steps, in a \textit{Maximize-Minimize} fashion. One step would consist in fixing the encoder's weights $\mathcal{W}$ and only minimizing \autoref{eq:penalty} w.r.t to the classifier's weights $\w$. At this step, $\ent(Y|\widehat{Z})$ would be fixed while $\widehat{Y}$ would be adjusted to minimize $\mathcal{D}_{KL}(Y||\widehat{Y}|\widehat{Z})$. Ideally, the KL term would vanish at the end of this step. In the following step, we would minimize \autoref{eq:penalty} w.r.t to the encoder's weights $\mathcal{W}$, while keeping the classifier fixed.
        \end{proof}

\section{Preliminary results with SPCE}\label{appendix:spce_experiment}

\begin{figure}[H]
    \centering
    \includegraphics[scale=0.45]{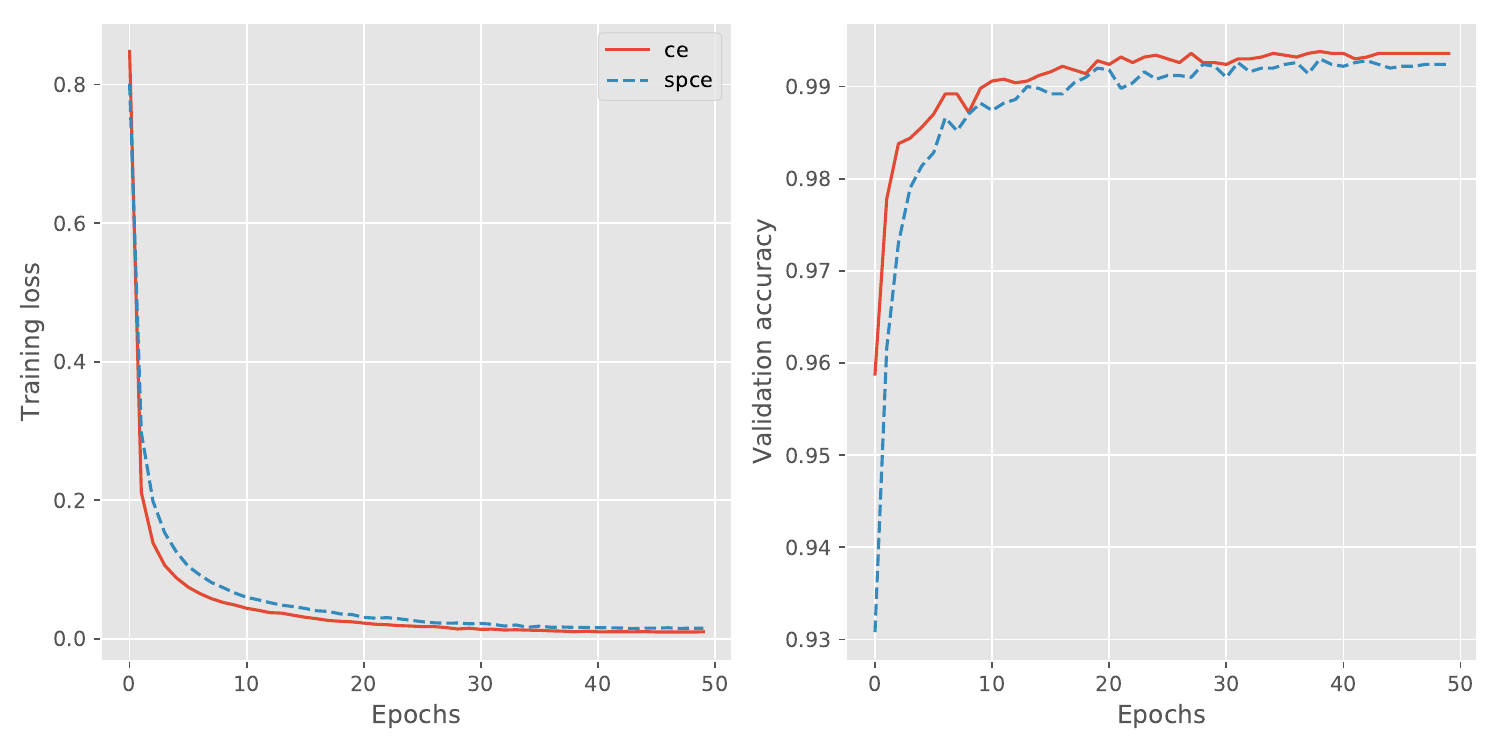}
    \caption{Evolution of the cross-entropy loss (CE) and the simplified pairwise cross-entropy (SPCE) during training on MNIST, as well as the validation accuracy for both losses.}
    \label{fig:CE_vs_PCE}
\end{figure}

In \autoref{fig:CE_vs_PCE}, we track the evolution of both loss functions and validation accuracy when training with $\mathcal{L}_{CE}$ and $\mathcal{L}_{SPCE}$ on MNIST dataset. We use a small CNN composed of four convolutional layers. The optimizer used is Adam. Batch size is set to 128, learning rate to $1e^{-4}$ with cosine annealing, weight decay to $1e^{-4}$ and feature dimension to $d=100$. \autoref{fig:CE_vs_PCE} supports the theoretical links that were drawn between Cross-Entropy and its simplied pairwise version SPCE. Particularly, this preliminary result demonstrates that SPCE is indeed employable as a loss, and exhibits a very similar behavior to the original cross-entropy. Both losses remain very close to each other throughout the training, and so remain the validation accuracies. 

\section{Analysis of ranking losses for Deep Metric Learning}

Some recent works \cite{cakir2019deep,wang2019ranked,rolinek2020optimizing} tackle the problem of deep metric learning using a rank-based approach. In other words, given a point in feature space $\z_i$, the pairwise losses studied throughout this work try to impose manual margins $m$, so that the distance between $\z_i$ and any negative point $\z_j^-$ is at least $m$. Rank-based losses rather encourage that all points are well ranked, distance-wise, such that $d(\z_i, \z_j^+) \leq d(\z_i, \z_j^-)$ for any positive and negative points $\z_j^+$ and $\z_j^-$. We show that our tightness/contrastive analysis also holds for such ranking losses. In particular, we analyse the loss proposed in \cite{cakir2019deep}. For any given query embedded point $\z_i$, let us call $D$ the random variable associated to the distance between $z_i$ and all other points in the embedded space, defined over all possible (discretized) distances $\mathcal{D}$. Furthermore, let us call $R$ the binary random variable that describes the relation to the current query point ($R^+$ and $R^-$ describe respectively a positive and negative relationship to $\z_i$). The loss maximized in \cite{cakir2019deep} reads:
\begin{align}
    \text{FastAP} = \sum_{d \in \mathcal{D}} \frac{P(D<d|R^+)P(R^+)}{P(D<d)}P(D=d|R^+)
\end{align}
Taking the logarithm, and using Jensen's inequality, we can lower bound this loss:
\begin{align}\label{eq:log_fastap}
    \log (\text{FastAP}) &\geq \sum_{d \in \mathcal{D}} P(D=d,R^+) \log(\frac{P(D<d|R^+)}{P(D<d)}) \nonumber \\
                         &= \underbrace{\displaystyle \mathop{\mathbb E}_{d \sim P(.,R^+)} \log P(D<d|R^+)}_{T_{AP}=\textsc{TIGHTNESS}} - \underbrace{\displaystyle \mathop{\mathbb E}_{d \sim P(.,R^+)} \log P(D<d)}_{C_{AP}=\textsc{CONTRASTIVE}}
\end{align}
To intuitively understand what those two terms are doing, let us imagine we approximate each of the expectations with a single point Monte-Carlo approximation. In other words, we sample a positive point $\z_j^{+}$, take its associated distance to $\z_i$, which we call $d^{+}$, then we approximate the tightness term as:
\begin{align}
    T_{AP} \approx \log P(D<d^{+}|R^+)
\end{align}
Maximizing $T_{AP}$ has a clear interpretation: it encourages all positive points to lie inside the hypersphere of radius $d^+$ around query point $\z_i$. Similarly:
\begin{align}
    C_{AP} \approx - \log P(D<d^{+})
\end{align}
Maximizing $C_{AP}$ also has a clear interpretation: it encourages all points (both positive and negative ones) to lie outside the hypersphere of radius $d^+$ around query point $\z_i$. Now, \autoref{eq:log_fastap} is nothing more than an expectation over all positive distance $d^+$ one could sample. Therefore, such loss can be analyzed through the same lens as other DML losses, i.e., one tightness term that encourages all points from the same class as $\z_i$ to lie close to it in the embedded space, and one contrastive term that oppositely refrains all points from approaching $\z_i$ closer than its current positive points.

\section{On the limitations of cross-entropy}

While we demonstrated that the cross-entropy loss could be competitive in comparison to pairwise losses, while being easier to optimize, there still exist scenarios for which a straightforward use of the CE loss becomes prohibitive. Hereafter, we describe two such scenarios. \\

\textbf{Case of relative labels: } The current setting assumes that absolute labels are given for each sample, \ie, each sample $\x_i$ belongs to a single absolute class $y_i$. However, DML can be applied to more general problems where the absolute class labels are not available. Instead, one has access to relative labels that only describe the relationships between points (\eg, a pair is similar or dissimilar). From these relative labels, one could still define absolute classes as sets of samples inside which every pair has a positive relationship. Note that with this definition, each sample may belong to multiple classes simultaneously, which makes the use of standard cross-entropy difficult.
However, with such re-formulation, our Simplified Pairwise Cross-Entropy (SPCE), which we hereby remind:
\begin{equation*}
    \lspce\tag{\ref{eq:spce}}
\end{equation*}
can handle such problems, just like any other pairwise loss. \\

\textbf{Case of large number of classes: } In some problems, the total number of classes K can grow to several millions. In such cases, even simply storing the weight matrix $\w \in 
\mathbb R^{K \times d}$ of the final classifier required by cross-entropy becomes prohibitive. Note that there exist heuristics to handle such problems with standard cross-entropy, such as sampling subsets of classes and solving those sub-problems instead, as was done in \cite{zhai2018classification}. However, we would be introducing new training heuristics (e.g., class sampling), which defeats the initial objective of using the cross-entropy loss. Again, the SPCE loss underlying the unary cross-entropy could again handle such cases, similarly to other pairwise losses, given that it doesn't require storing such weight matrix.